\documentclass{article}
\usepackage{xcolor}
\usepackage{microtype}
\usepackage{graphicx}
\usepackage{subcaption}
\usepackage{booktabs} 
\usepackage{multirow}
\usepackage{enumitem}
\usepackage[most]{tcolorbox}

\usepackage{hyperref}
\newcommand{\oursup}[1]{\textcolor{red}{(\,$\uparrow$ #1\,)}}

\usepackage[preprint]{icml2026}

\usepackage{amsmath}
\usepackage{amssymb}
\usepackage{mathtools}
\usepackage{amsthm}

\usepackage[capitalize,noabbrev]{cleveref}

\theoremstyle{plain}

\theoremstyle{definition}

\theoremstyle{remark}

\usepackage[textsize=tiny]{todonotes}

\begin{document}

\twocolumn[
  \icmltitle{SRVAU-R1: Enhancing Video Anomaly Understanding via Reflection-Aware Learning}

  \begin{icmlauthorlist}
    \icmlauthor{Zihao Zhao}{uiowa}
    \icmlauthor{Shengting Cao}{ualabama}
    \icmlauthor{Muchao Ye }{uiowa}
  \end{icmlauthorlist}
  
  \icmlaffiliation{uiowa}{Department of Computer Science, The University of Iowa, Iowa City, IA, USA}
  \icmlaffiliation{ualabama}{Department of Computer Science, Knox College,  Galesburg, IL, USA}

  \icmlcorrespondingauthor{Muchao Ye}{muchao-ye@uiowa.edu}

  \icmlkeywords{Machine Learning, ICML}

  \vskip 0.3in
]

\printAffiliationsAndNotice{} 
\begin{abstract}
Multi-modal large language models (MLLMs) have demonstrated significant progress in reasoning capabilities and shown promising effectiveness in video anomaly understanding (VAU) tasks. However, existing MLLM-based approaches remain largely focused on surface-level descriptions of anomalies, lacking deep reasoning over abnormal behaviors like explicit self-reflection and self-correction. To address that, we propose \textbf{S}elf-\textbf{R}eflection-Enhanced Reasoning for \textbf{V}ideo \textbf{A}nomaly \textbf{U}nderstanding (SRVAU-R1), a reflection-aware learning framework that incorporates reflection in MLLM reasoning. Specifically, SRVAU-R1 introduces the first reflection-oriented Chain-of-Thought dataset tailored for VAU, providing structured supervision with initial reasoning, self-reflection, and revised reasoning. Based on that, it includes a novel reflection-aware learning paradigm with supervised fine-tuning and reinforcement fine-tuning to enhance multi-modal reasoning for VAU. Extensive experiments on multiple video anomaly benchmarks demonstrate that SRVAU-R1 consistently outperforms existing methods, achieving significant improvements in both temporal anomaly localization accuracy and reasoning quality.\footnote{Data and codes will be made publicly available.}

\end{abstract}

\begin{figure}[t]
  \centering
  \includegraphics[width=0.5\textwidth]{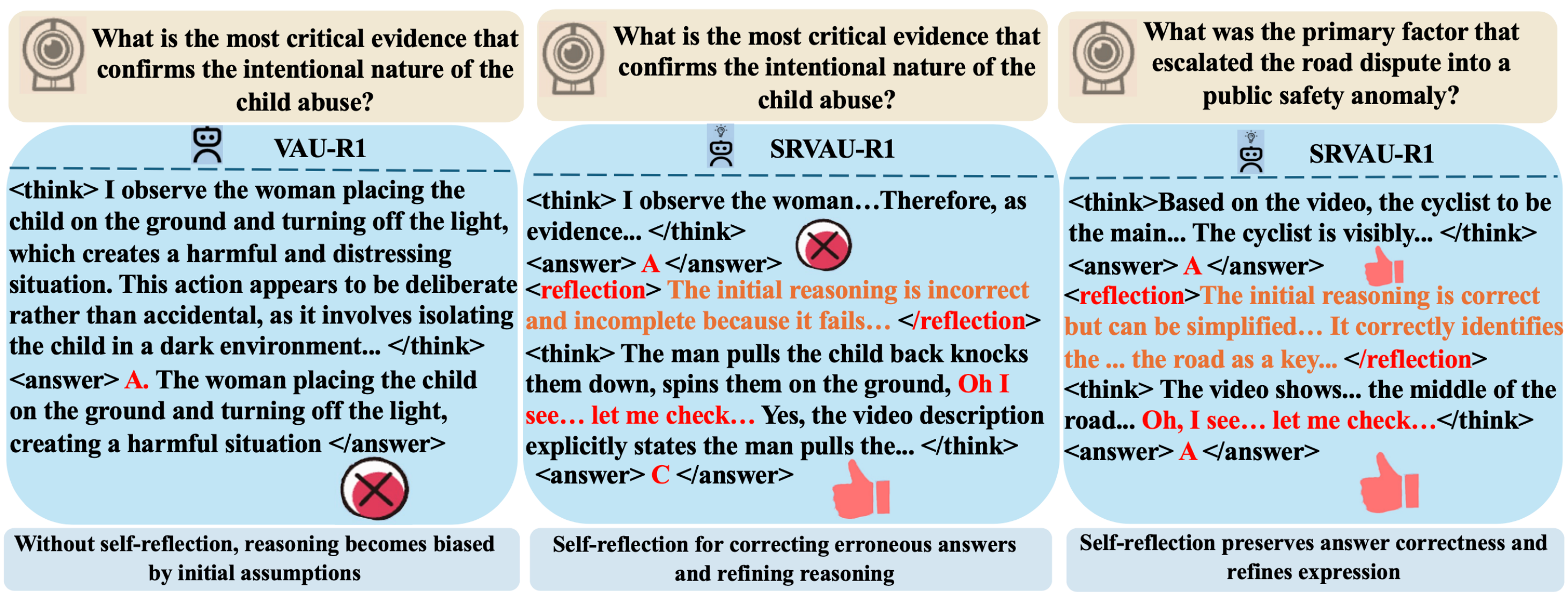}
  \caption{
  Existing VAU methods such as VAU-R1 provide only surface-level analysis and lack reflection on the generated results. In this paper, we propose a new method named SRVAU-R1 to include explicit self-reflection in VAU to correct or refine initial reasoning.
  }
  \label{fig:reflection_cases_and_results1.1}
\end{figure}
\section{Introduction}

Accurately identifying and deeply understanding anomalies such as traffic violations, theft, or sudden hazardous incidents from complex, unstructured video data remains a core challenge for intelligent perception systems in applications including public safety, traffic management, and urban governance \cite{ma2025sherlock,tang2024hawk,liu2019exploring}. Traditional methods primarily formulate this as the video anomaly detection task, localizing frame-level or clip-level anomalous temporal segments \cite{chen2023mgfn,ding2024learnable,gong2019memorizing,leng2024beyond,zhou2023dual,tian2021weakly}. Since the introduction of multi-modal large language models (MLLMs)~\cite{ye2025vera,doshi2023towards,zanella2024harnessing}, this task has transitioned into the video anomaly understanding (VAU)~\cite{zhu2025vau} task, aiming to generate accurate natural language descriptions for anomalous events. 

\noindent\textbf{Limitations.} However, existing MLLM-based approaches for VAU remain confined to surface-level anomaly descriptions \cite{du2024exploring,sultani2018real,yang2024follow}. The outstanding limitation is that they lack the ability to perform reflection over anomalous events, which is necessary in VAU to handle the ambiguity of anomalies. E.g., a video may be described as ``a person entering a restricted area'' by MLLMs, yet determining whether this event is truly anomalous requires reflection over temporal and contextual cues, such as whether the individual is a maintenance worker, responding to an emergency, or behaving maliciously. 

This assumption is fair for recent studies on multi-modal reasoning have demonstrated that explicit self-reflection and self-correction mechanisms play a critical role in enhancing the stability, consistency, and depth of model reasoning \cite{madaan2023self,kumar2024training,gandhi2025cognitive}. By reflecting on intermediate reasoning processes, models can identify and revise erroneous or redundant steps, resulting in more coherent reasoning chains with stronger causal interpretability \cite{zelazo2015executive}. Thus, it is imperative to enable MLLMs to reflect and think twice to correctly understand their causes, temporal evolution, and contextual semantics during conducting VAU. Thus, \emph{how to effectively integrate self-reflection mechanisms within multi-modal reasoning for VAU remains an open and challenging problem}. 

\noindent\textbf{Our Approach: SRVAU-R1.} This research question is nontrivial because of the lack of high-quality benchmarks and tailored learning paradigms that support fine-grained temporal reasoning in VAU for MLLMs. To address these challenges and introduce \textbf{s}elf-\textbf{r}eflection-enhanced reasoning for \textbf{VAU}, we propose a reflection-aware learning framework termed as SRVAU-R1. Specifically, SRVAU-R1 first includes a reflection-aware data construction pipeline, which polishes initial reasoning with reflection, augmenting training data with revised reasoning, as illustrated in \cref{fig:reflection_cases_and_results1.1}. Given that, SRVAU-R1 adopts a two-stage training strategy: In Stage 1, it conducts supervised fine-tuning (SFT)  on reflection-augmented training data that enable the model to acquire self-assessment and self-correction capabilities while learning anomaly reasoning, and in Stage 2, it utilizes reinforcement fine-tuning (RFT) built upon Group Relative Policy Optimization (GRPO) \cite{guo2025deepseek} to guiding the model to generate concise, coherent, and cognitively meaningful reflective reasoning processes. Particularly, we design a composite reward function that jointly optimizes reasoning correctness, reflection quality, and temporal consistency. 

Our main contributions can be summarized as follows:
\begin{itemize}[noitemsep,topsep=0pt]
    \item \textbf{Reflection-Oriented Data Construction Pipeline for VAU.} In SRVAU-R1, we propose a reflection-oriented data construction pipeline for VAU and build the first Chain-of-Thought (CoT) training dataset tailored for reflection-enhanced VAU. The dataset explicitly structures each sample into initial reasoning, self-reflection, and revised reasoning, providing crucial supervision signals for learning reasoning correction and self-evaluation in both SFT and RL.
    \item \textbf{Reflection-Aware Learning Framework for MLLMs in VAU.} SRVAU-R1 introduces a two-stage training framework for MLLMs in VAU that integrates reflection-enhanced supervised fine-tuning (SFT) with reflection-aware reinforcement learning (RL). This framework enables MLLMs not only to improve anomaly reasoning accuracy but also to diagnose and correct their own reasoning errors during inference, leading to more robust VAU.
    \item \textbf{Comprehensive empirical evaluation and analysis.} 
    We conduct extensive experiments and ablation studies across multiple VAU tasks and datasets. The results demonstrate that SRVAU-R1 consistently outperforms existing methods in anomaly understanding performance, temporal localization accuracy, and reasoning stability, clearly validating the effectiveness and necessity of explicit self-reflection mechanisms for complex video anomaly reasoning.
\end{itemize}

\section{Related Work}
\textbf{Reasoning Paradigms in Video Anomaly Tasks.}
The research paradigm on video anomaly has gradually shifted from detection-oriented formulations \cite{yao2023dota,huang2024long,liu2021hybrid,wu2020not,zhou2023dual,joo2023clip,liu2018future,lu2020few,ye2019anopcn} toward VAU \cite{lv2024video,zhu2025vau} due to the launch of MLLMs. 
Existing VAU methods primarily follow two lines of development: (1) prompt-driven approaches \cite{yang2024follow} which guide MLLMs to analyze anomalies through few-shot prompts or rule-based templates, offering low training cost but suffering from limited generalization, and (2) learning-driven approaches which internalize anomaly semantics and causal knowledge into model parameters through fine-tuning \cite{zhang2024holmes}. Compared to previous RL-based VAU methods like VAU-R1~\cite{zhu2025vau} and VAD-R1 \cite{huang2025vad}, SRVAU-R1 provides a novel learning pipeline based on SFT and RFT that empowers MLLMs in VAU to perform reflection-aware reasoning with a reflection-oriented data construction pipeline. Under SRVAU-R1, MLLMs will exhibit advanced reasoning with reflection, with improved reasoning quality. 

\textbf{RL for MLLM Reasoning.} 
In recent years, the emergence of high-capacity multi-modal models such as GPT-o3, Gemini 2.5, Seed1.5-VL, and DeepSeek-R1 has positioned RL as a key technique for improving reasoning capabilities during post-training of multi-modal models \cite{bi2025reasoning, feng2025video, huang2025vision,wang2024enhancing,zhou2025r1,guo2025deepseek}. Specifically, GRPO and later variants \cite{yue2025vapo, yu2025dapo,dou2025plan,liu2025understanding} have become mainstream RL methods for training LLMs due to optimization efficiency and reasoning stability. However, RL-based methods typically require substantial computational resources and large-scale training data. To reduce adaptation cost, several works combine SFT with RL, leveraging CoT cold-start initialization (e.g., Reason-RFT~\cite{tan2025reason} and Vision-R1~\cite{huang2025vision}) or more data-efficient visual RFT paradigms (e.g., Visual-RFT  \cite{liu2025visual}), and further extend these strategies to VAU methods~\cite{zhu2025vau,huang2025vad}. On the high level, our work shares the same vision as SRPO \cite{wan2025srpo}: we need to systematically model self-reflection and error correction across the SFT and RL stages. Compared to SRPO, SRVAU-R1 contains tailored reward design highlighting temporal localization.

\begin{figure*}[t]
  \centering
  \includegraphics[width=0.9\textwidth]{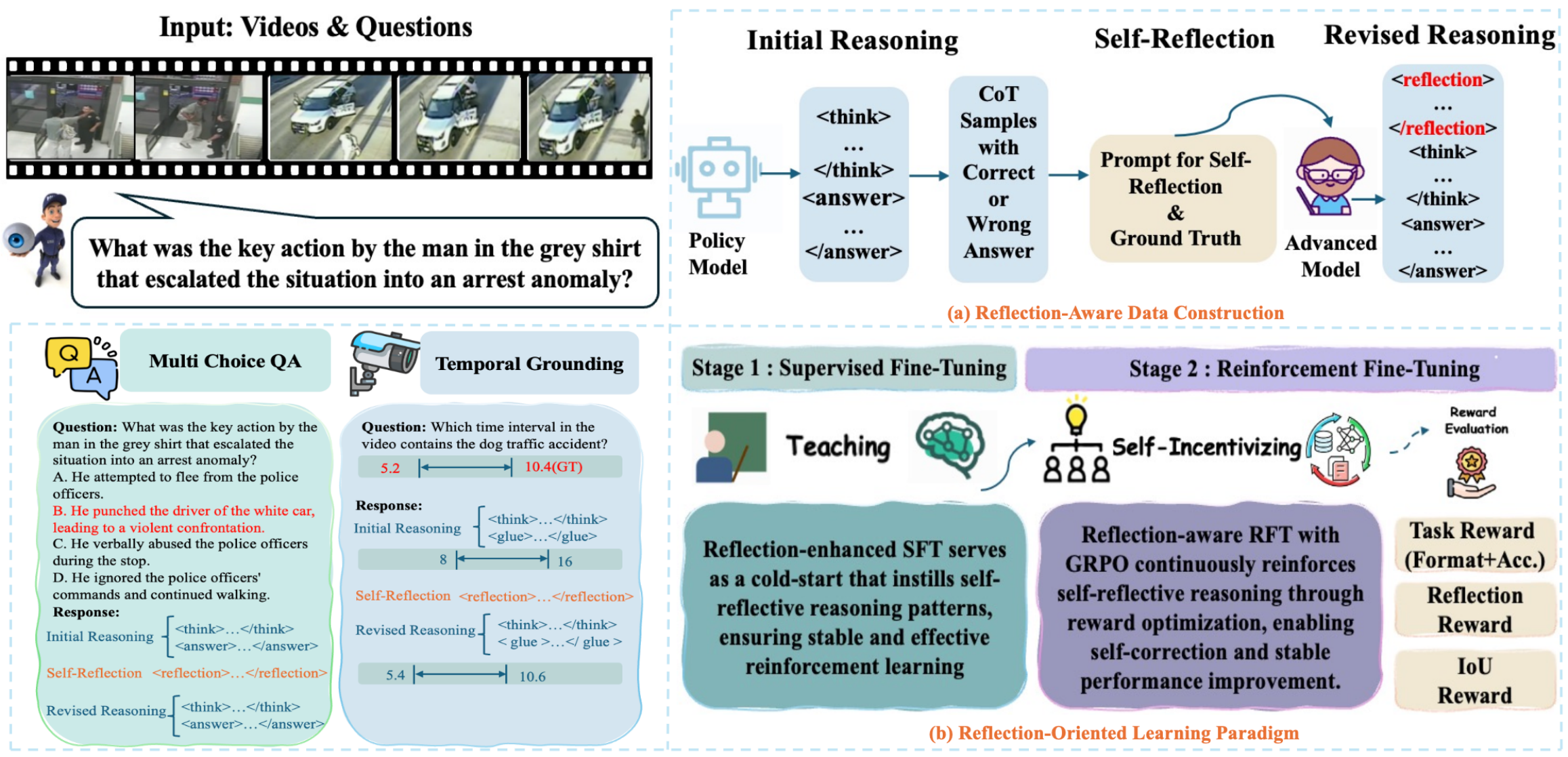}
  \caption{
  Overview of the SRVAU-R1 framework. First, SRVAU-R1 includes
  the (a) reflection-aware data construction pipeline, where initial chain-of-thought candidates and explicit self-reflection annotations are generated under the guidance of an advanced model.  SRVAU-R1 can be flexibly applied in VAU including multi-choice question answering and temporal anomaly grounding, which requires structured reasoning and reflection. Finally, SRVAU-R1 includes a novel (b) learning paradigm consisting of reflection-enhanced SFT and RFT. During the RL process, model parameters are optimized via task, reflection, and temporal IoU rewards.
  }
  \label{fig:srvau_overview}
\end{figure*}
\section{SRVAU-R1 Framework}
In this section, we present SRVAU-R1, a novel two-stage end-to-end training framework for VAU that places self-reflection at the core of multi-modal reasoning, as illustrated in \cref{fig:srvau_overview}. 
In Sec.~\ref{sec:data_construct}, we first present the reflection-aware data construction pipeline for crafting a reflection-oriented dataset to equip the model with fundamental capabilities for self-reflection in reasoning.
In Sec.~\ref{sec:reflect_paradigm}, we will introduce the learning paradigm of SRVAU-R1. 

\subsection{Reflection-Aware Data Construction}
\label{sec:data_construct}

In SRVAU-R1, we first need to prepare the learning of MLLMs with reflection-aware data to provoke reflection in their reasoning. This is necessary for such behavior is not explicitly modeled in the pretraining data of MLLMs. The construction of such datasets are illustrated in \cref{fig:srvau_overview}(a).

In this pipeline, we first employ Chain-of-Thought (CoT) prompting to obtain initial reasoning outputs from a base multi-modal model (e.g., \texttt{Qwen2.5-VL-3B}). Subsequently, the initial reasoning results, together with the ground-truth annotations, are fed into a higher-capacity multi-modal teacher model (e.g., \texttt{Qwen3-VL-30B}), which generates explicit reflections through two complementary pathways: correcting erroneous reasoning and refining correct but redundant reasoning. This dual-path reflection mechanism allows the model to simultaneously learn error diagnosis and reasoning optimization. The used prompt template is shown in Sec.~\ref{sec:prompt_template} in the Appendix  and some generated samples are shown in Sec.~\ref{sec:generation_sample} in the Appendix. 

Formally, given a VAU dataset $D=\{(v,q)\}$, given a question $q$ grounded on the video $v$, the base model $f$ first produces an \textbf{initial reasoning} output $a_1$. Subsequently, a high-capacity multi-modal model $f_{\rm teach}$ generates an explicit \textbf{self-reflection} signal $r$, which identifies potential reasoning errors such as evidence misinterpretation, temporal misalignment, or missing causal links. $f_{\rm teach}$ will provide the revised reasoning $a_2$. Through this process, we explicitly incorporate reasoning–reflection–revision structured supervision for VAU in learning.

\subsection{Reflection-Oriented Learning Paradigm}
\label{sec:reflect_paradigm}

In the learning paradigm of SRVAU-R1, as detailed in Sec.~\ref{sec:reflect_initial}, it first utilizes SFT to effectively expand the reasoning boundaries of MLLMs and enhance their cognitive capacity through the incorporation of external knowledge distillation. After that, in Sec.~\ref{sec:reflect_rl}, we introduce a reflection-aware RFT strategy built upon GRPO. By incorporating a carefully designed reflection-aware reward mechanism, the model is continuously encouraged to improve both the effectiveness of self-reflection and the temporal consistency of its reasoning during RL. 
Together, these two stages enable SRVAU-R1 to develop robust, interpretable, and temporally grounded reasoning behaviors for VAU.

\subsubsection{Cold-Start Initialization with SFT}
\label{sec:reflect_initial}
To achieve stable and effective reflection-aware VAU, we first perform a reflection-oriented cold-start initialization for the SRVAU-R1 policy model. This stage aims to inject fundamental self-reflection and self-correction capabilities into the model before RL to prevent potential unstable reflective behaviors caused by using RL only \cite{yue2025does}. 
 
Specifically, given the high-quality reflection-augmented training dataset $D_{\rm reflect} = \{(v,q,a_1,r,a_2)\}$ from Sec.~\ref{sec:data_construct}, we propose a reflection-oriented SFT strategy that explicitly injects reflection knowledge into the policy model \cite{shah2025rethinking}. Conditioned on this reflection, the policy model $f$ is optimized to generate a revised reasoning trajectory and the final answer $a_2$. As an SFT task, the training objective for the cold-start stage is the commonly used negative log-likelihood:
\vspace{-1.5em}
\begin{equation}
\begin{aligned}
\mathcal{L}_{\text{cold-start}}
= {} & - \mathbb{E}_{(q, a_1, r, a_2) \sim \mathcal{D}}
\Bigg[
\sum_{t=1}^{T}
\log \pi_{\text{initial}} \Big(
a_2 \mid q, a_1, \\
& \qquad \langle \text{reflection} \rangle r \langle / \text{reflection} \rangle
\Big)
\Bigg],
\end{aligned}
\end{equation}
where $r$ denotes the explicit reflection content and $a_2$ represents the reflection-corrected reasoning process together with the corresponding ground-truth answer. $\pi_{\text{initial}}$ is the policy (model parameters) of $f$.

This reflection-oriented cold-start initialization serves two complementary purposes: (1) By explicitly learning to revise initial reasoning under reflective guidance, the policy model gradually acquires the ability to identify reasoning errors, temporal inconsistencies, and missing causal structures, which are essential for robust VAU. (2) The reflection-enhanced cold start provides the model with structured reasoning priors, effectively preventing unstable or superficial reflection behaviors during subsequent reinforcement fine-tuning. As a result, reflection is progressively internalized as part of the model’s decision-making process rather than functioning as a post hoc correction.
Through this initialization, SRVAU-R1 smoothly transitions from general multi-modal understanding to reflection-aware video anomaly reasoning, laying a solid foundation for the subsequent reflection-aware RFT stage.

\subsubsection{Reflection-Aware RFT}
\label{sec:reflect_rl}
\noindent\textbf{Background.}
In the RFT stage, we adopt GRPO as the core optimization framework.
Specifically, given an input query $q$, the old policy $\pi_{\text{old}}$ samples a group of $G$ candidate responses $\{o_1, o_2, \ldots, o_G\}$. For each candidate response $o_i$, a task-specific reward $r_i$ is computed according to predefined reward functions. GRPO then normalizes these rewards within the group to compute the relative advantage for each candidate: 
$A_i
=
\frac{
r_i - \mathrm{mean}\!\left(\{r_1, r_2, \ldots, r_G\}\right)
}{
\mathrm{std}\!\left(\{r_1, r_2, \ldots, r_G\}\right)
}$.
Based on estimated advantages, the GRPO optimization objective is defined as: $\mathcal{L}_{\text{GRPO}}(\theta) = \mathbb{E}_{q,\{o_i\}\sim\pi_{\text{old}}} \left[ \frac{1}{G} \sum_{i=1}^{G} \min\left( \mathcal{J}_i, \widehat{\mathcal{J}}_i \right) - \beta \mathbb{D}_i \right]$, where $\mathcal{J}_i = \rho_i(\theta) A_i$, $ \widehat{\mathcal{J}}_i  = \text{clip}(\rho_i(\theta), 1-\epsilon, 1+\epsilon) A_i$, and $ \mathbb{D}_i = D_{\text{KL}}(\pi_\theta \| \pi_{\text{ref}})$, where $\epsilon$ is the clipping threshold that stabilizes policy updates, and $\beta$ is the coefficient of the KL-divergence regularization term, encouraging the learned policy to remain close to the frozen reference policy $\pi_{\text{ref}}$.

\noindent\textbf{Overall Reward Design in SRVAU-R1.}
Our reinforcement learning stage is built upon the SRPO~\cite{wan2025srpo} framework. Since the original reward formulation in GRPO primarily focuses on answer correctness and format adherence and lacks critical complex multi-modal reasoning abilities like reflection, SRPO ameliorates GRPO by including reflection-aware reward. However, the reward mechanism in SRPO does not fit in video understanding tasks and is only for static input tasks. This limitation is particularly evident in VAU, where models must reason over long temporal horizons with ambiguous visual evidence and evolving causal relationships.  To address this limit, we extend the SRPO reward mechanism with a temporal IoU reward term.  Under this reward mechanism, this enhancement optimizes the correctness of the final output, the ability to self-evaluate, as well as the capacity of understanding temporal inputs throughout the reasoning process.
For each candidate response, the total reward is defined as
\begin{equation}
R_{\text{total}} = \alpha R_{\text{task}} + \beta R_{\text{reflection}} + \gamma R_{\text{tIoU}},
\end{equation}
where $R_{\text{task}}$ denotes the task-level reward, $R_{\text{reflection}}$ measures the quality and effectiveness of self-reflection, and $R_{\text{tIoU}}$ evaluates temporal localization accuracy. The coefficients $\alpha$, $\beta$, and $\gamma$ balance the contributions of each component.

\textbf{Task Reward.}
The task reward enforces basic correctness and structural consistency of the model output. It consists of a format reward and an accuracy reward:
\begin{equation}
R_{\text{task}} = R_{\text{format}} + R_{\text{accuracy}}.
\end{equation}
The format reward $R_{\text{format}}$ encourages the model to follow a standardized
reasoning structure (e.g., using $\langle \texttt{think} \rangle \cdots
\langle / \texttt{think} \rangle$ tags), thereby ensuring the consistency and
parsability of the generated reasoning process. We set $R_{\text{format}} = 0.5$
if the output format is correct, and $R_{\text{format}} = 0$ otherwise.
The accuracy reward $R_{\text{accuracy}}$ evaluates whether the model's initial
reasoning output matches the ground-truth label. We set $R_{\text{accuracy}} = 0.5$
if the first prediction is correct, and $R_{\text{accuracy}} = 0$ otherwise.
This design ensures that the model remains aligned with the task objective and avoids relying solely on post-hoc reflection.

\textbf{Reflection Reward.}
The reflection reward explicitly evaluates how effectively the model refines its reasoning through self-assessment and correction. It is defined as:
\begin{equation}
R_{\text{reflection}} = I_{\text{eff}} + I_{\text{ref}} + \alpha f_{\text{len}}(L_{\text{response}}),
\end{equation}
\noindent where $I_{\text{ref}} = 0.25$ if the reflection segment is correctly formatted using the \texttt{<reflect>} tag, and $I_{\text{ref}} = 0$  if it is incorrectly formatted;
$f_{\text{len}}(L_{\text{response}})$ is a reflection brevity regularization term that promotes concise yet informative reflections: $f_{\text{len}}(L_{\text{response}}) = \exp \left( -\frac{|L_{\text{response}} - T_{\text{target}}|}{T_{\text{max}} - T_{\text{target}}} \right)$ which softly penalizes overly verbose reflections and stabilizes gradient behavior; and 
    \begin{equation}
    I_{\text{eff}}=
    \begin{cases}
    +0.25, & \text{C}\!\rightarrow\!\text{C},\\
    +0.5,  & \text{W}\!\rightarrow\!\text{C},\\
    0,     & \text{W}\!\rightarrow\!\text{W},\\
    -0.25, & \text{C}\!\rightarrow\!\text{W},
    \end{cases}
    \end{equation}
which measures the actual effectiveness of reflection in improving reasoning outcomes. Here, \text{C} and \text{W} denote whether the prediction is Correct or Wrong before and after applying reflection, respectively.
This formulation explicitly distinguishes effective reflection from superficial or harmful reflection, preventing the model from exploiting reflection through verbosity or structural compliance alone.

\textbf{Temporal IoU Reward.}
To provide fine-grained supervision for temporal localization, we incorporate a temporal overlap reward based on Intersection-over-Union (IoU). Rather than serving as a standalone localization objective, this reward functions as a complementary signal that aligns temporal reasoning outcomes with ground-truth anomaly spans. Formally, the temporal reward is defined as
\begin{equation}
R_{\text{tIoU}} =
\begin{cases}
1, & \text{normal and correct},\\
\mathrm{IoU}\!\big([s_1,s_2],[s_1^\ast,s_2^\ast]\big), & \text{anomaly and correct},\\
0, & \text{otherwise}.
\end{cases}
\end{equation}

\subsection{Aha Moment in SRVAU-R1}
We observe that SRVAU-R1 exhibits explicit and consistent ``aha moments'' \cite{guo2025deepseek} during video anomaly reasoning (We will illustrate this with qualitative results in Sec.~\ref{sec:case}). When confronted with complex scenarios involving ambiguous temporal cues, unclear anomaly boundaries, or multi-step causal inference, the model’s initial reasoning is often formed based on partial or local evidence. Subsequently, triggered by the reflection signal, the model actively revisits the video context and re-evaluates its previous conclusions, leading to a revision of the reasoning trajectory \cite{pan2025unlocking}. This behavior demonstrates that the model is able to self-reflect: 
identifying evidence omissions or causal inconsistencies in its initial reasoning and abruptly revising its judgment during the second reasoning stage, where it re-localizes the critical temporal window of the anomaly or corrects its core causal hypothesis.
\begin{table*}[t]
  \caption{Comparison results of QA accuracy and VAU-Eval results on MSAD and UCF-Crime datasets. Due to the space limit, the comparison on ECVA is shown in Sec.~\ref{sec:ecva} in the Appendix.}
  \vspace{-0.1in}
  \label{tab:qa_task}
  \begin{center}
    \begin{small}
        \setlength{\tabcolsep}{1.5pt}
        \renewcommand{\arraystretch}{1.05}
        \begin{tabular}{c l | cc | cccccc}
          \toprule
          \multirow{2}{*}{Dataset} & \multirow{2}{*}{Model}
          & \multicolumn{2}{c|}{QA Accuracy}
          & \multicolumn{6}{c}{VAU-Eval} \\
          \cmidrule(lr){3-4}\cmidrule(lr){5-10}
          & & Acc$_{\mathrm{w/o\ think}}$ & Acc$_{\mathrm{w/\ think}}$
          & CLS$\uparrow$ & KM$\uparrow$ & FLU$\uparrow$ & INF$\uparrow$ & FAC$\uparrow$ & Total$\uparrow$ \\
          \midrule

          \multirow{8}{*}{MSAD}
          & InternVL2.5-2B
          & 76.67 & 72.08 & 6.84 & 6.23 & 8.55 & 6.64 & 6.64 & 34.90 \\
          & Qwen2.5-VL-7B
          & 84.58 & 83.33 & 6.75 & 6.41 & 9.27 & 7.74 & 6.92 & 37.08 \\
          & InternVL2.5-8B-MPO
          & 82.50 & 84.17 & 6.83 & 6.33 & 8.32 & 6.37 & 6.86 & 34.72 \\
          \cmidrule(l){2-10}
          & Qwen2-VL-2B
          & 77.08 & 72.50 & 5.94 & 5.43 & 8.77 & 6.29 & 5.90 & 32.25 \\
          & Qwen2.5-VL-3B
          & 85.83 & 82.50 & 5.77 & 5.24 & 9.02 & 6.74 & 5.70 & 32.47 \\
          & HolmesVAU
          & 85.00 & 86.25 & 3.73 & 2.72 & 6.82 & 3.55 & 3.33 & 20.15 \\
          & VAU-R1
          & 88.33 & 87.08 & 5.97 & 5.49 & 9.05 & 6.84 & 6.03 & 33.38 \\
          & \textbf{SRVAU-R1}
          & \textbf{89.58} \oursup{3.75} & \textbf{91.25} \oursup{8.75}
          & 7.65 & 6.98 & 9.78 & 7.08 & 7.43 & \textbf{38.30} \\
          \midrule

          \multirow{8}{*}{UCF-Crime}
          & InternVL2.5-2B
          & 84.86 & 68.13 & 4.40 & 3.08 & 8.09 & 5.69 & 3.47 & 24.74 \\
          & Qwen2.5-VL-7B
          & 92.03 & 89.64 & 4.80 & 3.73 & 8.95 & 7.05 & 4.25 & 28.78 \\
          & InternVL2.5-8B-MPO
          & 89.64 & 90.44 & 3.79 & 3.20 & 8.23 & 5.77 & 3.48 & 24.47 \\
          \cmidrule(l){2-10}
          & Qwen2-VL-2B
          & 87.25 & 83.67 & 3.47 & 2.48 & 7.75 & 4.49 & 2.82 & 21.02 \\
          & Qwen2.5-VL-3B
          & 91.63 & 83.27 & 4.31 & 2.88 & 8.70 & 5.95 & 3.27 & 25.10 \\
          & HolmesVAU
          & 86.45 & 85.66 & 3.05 & 1.97 & 6.30 & 3.08 & 2.39 & 16.79 \\
          & VAU-R1 
          & 92.03 & 91.63 & 4.42 & 2.98 & 8.71 & 5.89 & 3.39 & 25.49 \\
          & \textbf{SRVAU-R1}
          & \textbf{92.82} \oursup{1.19} & \textbf{96.81} \oursup{13.54}
          & 7.22 & 6.76 & 9.16 & 6.56 & 6.96 & \textbf{36.66} \\
          \bottomrule
        \end{tabular}
    \end{small}
  \end{center}

    \vspace{-0.1in}
\end{table*}
\section{ Experiments }

\subsection{Experimental Settings}
\textbf{Training Data.} 
Based on VAU-Bench~\cite{zhu2025vau}, we select samples from three public video anomaly datasets, including UCF-Crime, ECVA, and MSAD, as the raw data sources, and systematically filter and restructure them following the data construction pipeline described in Sec.~\ref{sec:data_construct}. This process results in a training dataset integrated with reflection-enhanced annotations, consisting of 4,602 video samples, covering 19 major anomaly categories, including violent behaviors, accidents, hazardous activities, and sudden anomalous events commonly observed in real-world scenarios. The resulting dataset exhibits substantial diversity in video duration, anomaly patterns, and semantic complexity, providing strong support for reflection-aware anomaly reasoning. More details regarding the dataset construction and annotation process are provided in Sec.~\ref{app:data} in the Appendix.

\textbf{Metrics and Baselines.} We evaluate our model on VAU-Bench~\cite{zhu2025vau}. Following their settings, for the question answering (QA) task, we adopt multiple-choice accuracy as the evaluation metric; for temporal anomaly localization, we report the temporal mean Intersection-over-Union (mIoU) as well as recall at different IoU thresholds (R@0.3, R@0.5, and R@0.7) to comprehensively assess localization performance; and for generated reasoning and explanatory outputs, we use VAU-Eval in~\cite{zhu2025vau} and employ DeepSeek-V3 \cite{liu2024deepseek} as the judge model and evaluate each response along five dimensions: classification correctness, key concept alignment, fluency, informativeness, and factual consistency. Each dimension is scored on a 10-point scale, enabling fine-grained assessment of reasoning quality. Following~\cite{zhu2025vau}, we use frozen models from Qwen-VL and InternVL families as baselines. We also compare our results with VAU-R1 and HolmesVAU~\cite{zhang2025holmes} for evaluating the effectiveness of our learning paradigm.

\textbf{Implementation Setup.} Our main experiments are conducted using the Qwen2.5-VL-3B-Instruct \cite{bai2025qwen2} models as the base model $f$ and  Qwen3-VL-30B-Instruct \cite{yang2025qwen3} as the teacher model $f_{\rm teach}$. The used prompt template for initial reasoning, self-reflection, and revised reasoning is shown in Sec.~\ref{sec:prompt_template} in the Appendix. Both the self-reflection cold-start SFT stage and the subsequent reinforcement learning stage are trained using 4 NVIDIA RTX PRO 6000 GPUs. All experiments employ full-parameter fine-tuning, without introducing any parameter-efficient adapters or LoRA modules. During inference, for Qwen-VL–based models, video frames are sampled at 1 FPS. For InternVL-based models which are used as a teacher model, we uniformly sample 16 frames per video as input. This unified inference protocol ensures fair comparison across different model families.

\subsection{Comparison Results}
\textbf{Evaluation on QA-Guided Reasoning.} As shown in Table \ref{tab:qa_task}, the experimental results demonstrate that SRVAU-R1 consistently and significantly outperforms open-source general-purpose multi-modal baseline models across all evaluated datasets. On both MSAD and UCF-Crime, SRVAU-R1 achieves the best performance in terms of multiple-choice QA accuracy and Total (aggregated score over five dimensions) in VAU-Eval, exhibiting clear advantages over widely used MLLM baselines and VAU methods. This observation indicates that the observed performance gains do not stem from increased model capacity or supervised fine-tuning alone, but rather from the effective integration of reflection-aware training mechanisms. 

In addition, we observe that baselines often perform worse when explicit reasoning generation is enabled (i.e. $\mathrm{Acc}_{\text{w/think}}$) compared to when reasoning is disabled (i.e. $\mathrm{Acc}_{\text{w/o think}}$). This phenomenon suggests that naive CoT prompting may introduce reasoning hallucinations, which in turn degrade final answer accuracy. In contrast, SRVAU-R1 effectively mitigates this issue by explicitly modeling self-reflection during both SFT and RFT stages. Overall, these results strongly support our central claim: explicitly enhancing a model’s self-reflection capability during both supervised and reinforcement fine-tuning substantially improves high-level semantic reasoning, decision accuracy, and reasoning stability in VAU, enabling a shift from mere anomaly recognition toward deeper anomaly understanding.

\begin{table}[t]
  \caption{Comparison results of temporal anomaly grounding performance on MSAD and UCF-Crime datasets.}
        \vspace{-0.1in}
  \label{tab:temporal_grounding}
  \begin{center}
    \begin{small}
      \setlength{\tabcolsep}{1.5pt}
      \renewcommand{\arraystretch}{1.05}
      \begin{tabular}{l l cccc}
        \toprule
        Dataset & Model & mIoU & R@0.3 & R@0.5 & R@0.7 \\
        \midrule

        \multirow{4}{*}{MSAD}
        & Qwen2-VL-2B     & 0.00 & 0.00 & 0.00 & 0.00 \\
        & Qwen2.5-VL-7B   & 17.57 & 26.67 & 11.67 & 3.33 \\
        & Qwen2.5-VL-3B   & 11.05 & 15.42 & 5.35 & 1.67 \\
        & \textbf{SRVAU-R1}
        & \textbf{20.40} \oursup{9.35} & 30.00 & 19.17 & 8.33 \\
        \midrule

        \multirow{4}{*}{ECVA}
        & Qwen2-VL-2B     & 0.17 & 0.30 & 0.00 & 0.00 \\
        & Qwen2.5-VL-7B   & 5.71 & 7.96 & 4.73 & 2.99 \\
        & Qwen2.5-VL-3B   & 6.35 & 7.21 & 1.99 & 0.50 \\
        & \textbf{SRVAU-R1}
        & \textbf{44.42} \oursup{38.07} & 63.68 & 42.29 & 24.13 \\
        \bottomrule
      \end{tabular}
    \end{small}
  \end{center}
        \vspace{-0.2in}
\end{table}

\textbf{Evaluation on Temporal Anomaly Grounding.} Beyond QA-based anomaly reasoning, we further evaluate the model’s performance on Temporal Anomaly Grounding (TAG) to assess its temporal generalization under out-of-distribution (OOD) settings. Notably, in this experimental setup, the model is trained exclusively on the ECVA dataset, while MSAD is entirely reserved as an OOD test set, thereby avoiding any direct fitting to the test distribution. As reported in Table \ref{tab:temporal_grounding}, SRVAU-R1 consistently outperforms all open-source general-purpose multi-modal baselines on both MSAD and ECVA, achieving the best results across mIoU and recall at different IoU thresholds. In particular, on the OOD MSAD dataset, SRVAU-R1 attains an mIoU of 20.40, representing a 9.35-point improvement over Qwen2.5-VL-3B, which demonstrates robust temporal localization capability under unseen data distributions. On ECVA, the advantage of SRVAU-R1 becomes even more pronounced, with an mIoU of 44.42, outperforming the strongest baseline by 38.07 points, and maintaining a clear lead under stricter localization criteria such as R@0.7. These results indicate that reflection-enhanced reasoning not only facilitates precise modeling of semantic anomaly boundaries but also improves global temporal consistency in anomaly localization. Taken together, these findings show that introducing reflection-aware mechanisms during RFT significantly enhances the model’s generalization ability for TAG under unseen distributions. Compared to baselines relying solely on SFT or direct inference, reflection-enhanced reasoning equips the model with self-correction capability during temporal decision-making, resulting in greater stability and robustness  in complex, cross-distribution VAU scenarios.\\

\begin{table}[t]
  \caption{Influence of reflection data on video anomaly understanding.}
  \label{tab:ablation_reflection_data}
  \centering
  \small
  \setlength{\tabcolsep}{10pt}
  \renewcommand{\arraystretch}{1.15}
  \begin{tabular}{l cc}
    \toprule
    Method & UCF-Crime & MSAD \\
    \midrule
    Qwen2.5-VL-3B              & 83.27 & 82.50 \\
    +GRPO(w/o Reflection Data) & 91.63 & 87.08 \\
    \midrule
    \textbf{SRVAU-R1 (Ours)}   & \textbf{96.81} & \textbf{91.25} \\
    \bottomrule
  \end{tabular}
\end{table}

\subsection{Ablation Studies} We further investigate the impact of reflection-enhanced data construction, training-stage design, and the reflection reasoning paradigm itself in SRVAU-R1 on both reasoning quality. Without loss of generality, all ablation experiments are conducted under the default setting with Qwen2.5-VL-3B as $f$ in QA datasets.
\textbf{Effect of Reflection Data (w/o Reflection Data vs. SRVAU-R1).} As shown in \cref{tab:ablation_reflection_data}, we first remove the reflection-enhanced data construction process and train the model using only the original anomaly annotations with standard CoT reasoning. The results show a substantial degradation in temporal localization accuracy, reasoning coherence, and prediction stability when explicit reflection supervision is absent. This finding indicates that conventional reasoning signals alone are insufficient for complex VAU. Reflection-augmented data plays a critical role in guiding the model toward self-correction and causal reasoning, enabling it to revise erroneous inference paths and better align predictions with high-level anomaly semantics.

\begin{table}[t]
  \caption{Influence of isolating reflection-aware SFT and reflection-aware RL.}
  \label{tab:ablation_sft_rl}
  \centering
  \small
  \setlength{\tabcolsep}{8pt}
  \renewcommand{\arraystretch}{1.15}
  \begin{tabular}{l cc}
    \toprule
    Method & UCF-Crime & MSAD \\
    \midrule
    Qwen2.5-VL-3B              & 83.27 & 82.50 \\
    Reflection-SFT only         & 96.02 & 90.83 \\
    Reflection-RL only          & 93.23 & 88.48 \\
    \midrule
    \textbf{SRVAU-R1 (Ours)}    & \textbf{96.81} & \textbf{91.25} \\
    \bottomrule
  \end{tabular}
\end{table}

\textbf{Isolating the Roles of Reflection SFT and Reflection RL.} As presented in \cref{tab:ablation_sft_rl}, to disentangle the respective contributions of reflection mechanisms in  SFT and RFT, we evaluate the following two ablation settings: (1) applying reflection-enhanced data during SFT, without reflection-aware RL, and (2) introducing reflection-aware rewards during RL, without reflection-enhanced SFT initialization.
The results indicate that reflection-enhanced SFT alone improves reasoning quality to some extent, but remains insufficient for handling complex temporal decision-making and out-of-distribution (OOD) scenarios. Conversely, introducing reflection-aware rewards only during RL fails to reliably elicit stable reflection behavior, leading to training instability and slower convergence. In contrast, SRVAU-R1 benefits from a reflection-enhanced SFT stage that provides a well-structured cold-start initialization, followed by reflection-aware RL that continuously reinforces reflective behavior. This two-stage synergy yields consistently superior performance over both single-stage alternatives, demonstrating that effective reflection learning critically depends on the coordination between SFT and RL.

\begin{table}[t]
  \caption{Influence of reflection data scale on the effectiveness of SRVAU-R1.}
  \label{tab:ablation_reflection_config}
  \centering
  \small
  \setlength{\tabcolsep}{5pt}
  \renewcommand{\arraystretch}{1.15}
  \begin{tabular}{l c cc}
    \toprule
    Method &RL Data Size & UCF-Crime & MSAD \\
    \midrule
    GRPO                           & 1.5K &89.24 & 85.41\\
    GRPO                           & 3k & 91.63 & 87.08 \\
    SRVAU-R1                       & 1.5k & 94.42 & 87.50 \\
    \textbf{SRVAU-R1 (Ours)}       & 3k & \textbf{96.81} & \textbf{91.25} \\
    \bottomrule
  \end{tabular}
\end{table}

\textbf{Impact of Reflection Data Scale (Partial Reflection Dataset vs. SRVAU-R1)}. As shown in Table~\cref{tab:ablation_reflection_config}, we further examine the effect of reflection data scale by training SRVAU-R1 with randomly sampled subsets of the reflection dataset. As the size of the reflection data decreases, model performance degrades progressively. Nevertheless, even under severely reduced reflection data settings, SRVAU-R1 still outperforms baselines that entirely lack reflection mechanisms. This result suggests that reflection-enhanced training is data-efficient, while also highlighting that sufficient reflection data is important for fully realizing its potential. 
\begin{table}[t]
  \caption{Influence of the reflection paradigm (Two-Step Thinking vs. Think--Reflect--Rethink) and teacher models during reflection-aware reasoning .}
  \label{tab:reflection_paradigm}
  \centering
  \small
  \setlength{\tabcolsep}{5pt}
  \renewcommand{\arraystretch}{1.15}
  \begin{tabular}{l l cc}
    \toprule
    Method& UCF-Crime & MSAD \\
    \midrule
    Two-Step Thinking  & 89.64 & 85.41 \\
    SRVAU-R1 (InternVL3-38B Teacher) & 95.62 & 88.33 \\
    \midrule
    \textbf{SRVAU-R1 (Ours)}
      & \textbf{96.81} & \textbf{91.25} \\
    \bottomrule
  \end{tabular}
\end{table}

\textbf{Reflection Data from Alternative Teacher Models (Alternative Reflection Teacher vs. SRVAU-R1).} As shown in \cref{tab:reflection_paradigm}, to verify that the observed gains are not attributable to a specific teacher model, we generate reflection data using different multi-modal large models and retrain SRVAU-R1 accordingly.  Specifically, we use InternVL3-38B as $f_{\rm teach}$ to craft reflection-aware data $D_{\rm reflect}$ for SFT and RFT in SRVAU-R1. Despite variations in linguistic style and descriptive granularity across teachers, SRVAU-R1 exhibits consistent performance improvements. This observation indicates that the performance gains primarily arise from the reflection structure itself, rather than from distillation effects tied to a particular teacher model, further demonstrating the robustness and generality of the proposed approach.

\textbf{Effectiveness of the Reflection Paradigm (Two-Step Thinking vs. Think–Reflect–Rethink).}\quad Finally, to validate the necessity of the proposed ``Think–Reflect–Rethink'' reasoning paradigm, we compare SRVAU-R1 with a GRPO-based Two-Step Thinking baseline. In this alternative setting, the model generates two consecutive \texttt{<think>} and \texttt{</think>} reasoning steps and is trained using task rewards and consistency-based relational rewards, without introducing explicit reflection markers. The results in Table \cref{tab:reflection_paradigm} show that Two-Step Thinking fails to provide consistent improvements over vanilla GRPO, yielding only marginal gains on a small subset of datasets. In contrast, SRVAU-R1 significantly improves both anomaly reasoning accuracy and temporal localization performance by explicitly reflecting on and correcting prior reasoning. This comparison clearly demonstrates that simply increasing the number of reasoning steps cannot substitute for explicit self-reflection mechanisms. Reflection plays an indispensable role in guiding the model to identify and revise flawed reasoning trajectories, thereby enabling more reliable and interpretable VAU.

\begin{figure}[t]
  \centering
  \includegraphics[width=0.45\textwidth]{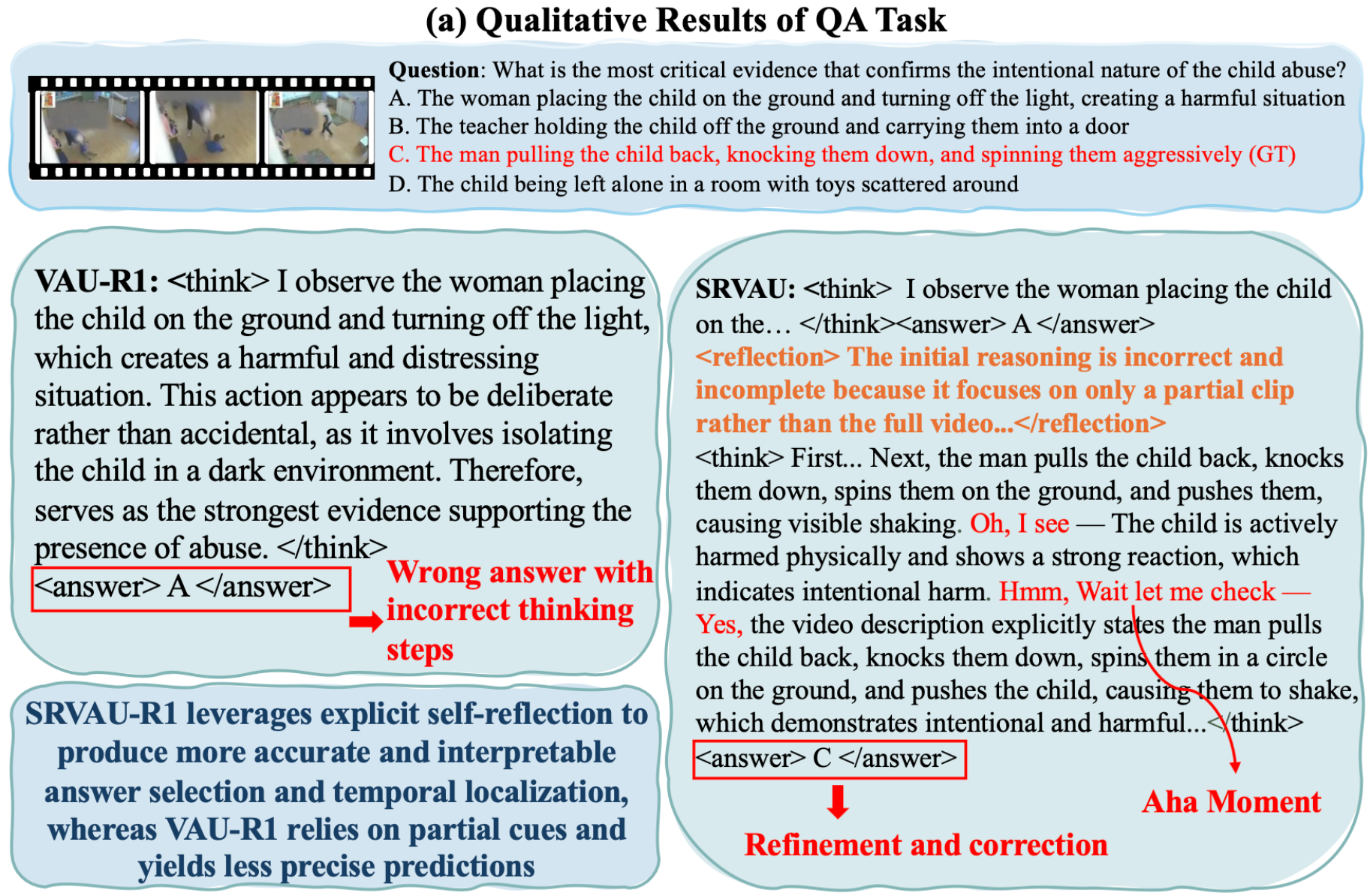}
  \vspace{0.15in}
  \includegraphics[width=0.45\textwidth]{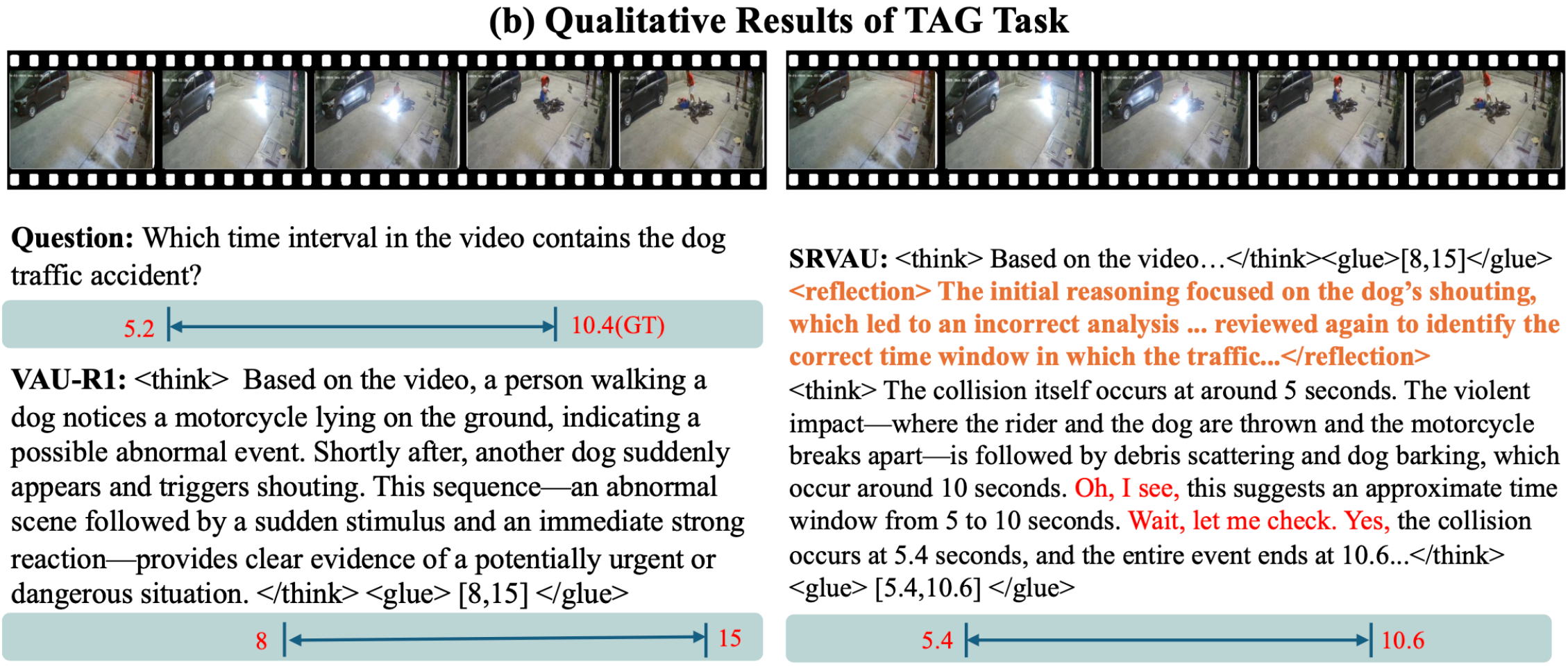}
  \caption{
  Qualitative results of \textbf{(a) QA} and \textbf{(b) TAG} tasks. All ground truth answers and anomaly intervals are highlighted in red. VAU-R1 and SRVAU-R1 reason under the same CoT prompting. SRVAU-R1 leverages explicit self-reflection to produce more accurate and interpretable answer selection and temporal localization, whereas VAU-R1 relies on partial cues and yields less precise predictions.
  }
  \label{fig:qualitative_reflection_cases}
\end{figure}

\subsection{Qualitative Results and Case Studies}
\label{sec:case}
We present qualitative case studies on two different VAU tasks, comparing
VAU-R1  and our SRVAU-R1 under the same  CoT prompt. In the  QA task shown in \cref{fig:qualitative_reflection_cases}(a), VAU-R1 relies excessively on local visual cues during its initial reasoning and consequently selects a benign interpretation of the observed behavior. In contrast, SRVAU-R1 explicitly recognizes the incompleteness of its initial reasoning during the reflection stage. By revisiting the full video context, it correctly identifies the man’s aggressive actions as the decisive evidence of intentional abuse. This correction is accompanied by a clear ``aha moment'', in which the model abandons its prior hypothesis and fundamentally revises its reasoning trajectory. In the TAG task shown in \cref{fig:qualitative_reflection_cases}(b), VAU-R1 produces a coarse anomaly interval that is temporally misaligned and lacks a clear causal explanation. By comparison, SRVAU-R1 becomes aware during reflection that its initial analysis overemphasizes non-essential cues. It then re-examines the temporal structure of the video and accurately localizes the anomaly interval to 5.4s–10.6s. Together, these two independent case studies demonstrate that SRVAU-R1, through explicit self-reflection, enables models to self-correct flawed initial reasoning, leading to more accurate, stable, and interpretable reasoning behaviors in both semantic understanding and temporal localization.

\section{Conclusions}
In this paper, we propose SRVAU-R1, a reflection-aware data construction and learning framework for  MLLMs in VAU.
Under this pipeline, we can construct a high-quality reflection-augmented dataset that provides structured supervision in the form of initial reasoning, explicit self-reflection, and revised reasoning. Given that, by combining reflection-enhanced SFT with reflection-aware RL, SRVAU-R1 not only improves reasoning accuracy but also allows the model to reflect and correct its own reasoning errors during inference, leading to more robust reasoning behaviors. Extensive experiments across multiple video anomaly benchmarks demonstrate that SRVAU-R1 consistently improves both anomaly understanding performance and reasoning stability, validating the effectiveness of reflection-aware training strategies for complex video reasoning tasks.

\section*{Impact Statement}
This work proposes SRVAU-R1, a reflection-aware learning framework for video anomaly understanding, with the goal of improving the stability of multimodal large language model reasoning in complex video scenarios.  Models with more reliable anomaly understanding capabilities have the potential to support safety-related applications such as public surveillance, abnormal behavior alerting, and incident analysis, thereby assisting human decision-making processes.  At the same time, video anomaly scenarios may involve violent, dangerous, or chaotic behaviors.  We strictly adhere to established ethical guidelines, and all data used in this study are drawn from publicly available datasets and are employed solely for academic research in accordance with the original publishers’ terms of use.  Overall, SRVAU-R1 aims to advance research in video anomaly understanding toward more reliable and interpretable reasoning.  Future work should continue to address robustness, fairness, and privacy considerations to ensure responsible deployment in real-world settings.

\bibliography{references}
\bibliographystyle{icml2026}

\newpage
\appendix
\onecolumn
\section{Training Dataset}
\label{app:data}
\begin{figure*}[t]
  \centering
  \begin{subfigure}[t]{0.45\textwidth}
    \centering
    \includegraphics[width=0.9\textwidth]{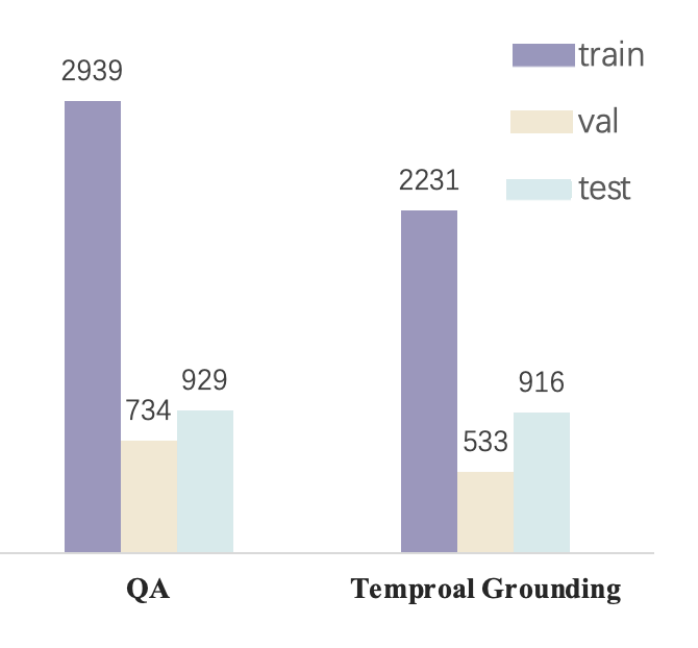}
    \caption{Dataset Split Statistics Across Tasks.}
    \label{fig:dataset_split}
  \end{subfigure}
  \hfill
  \begin{subfigure}[t]{0.54\textwidth}
    \centering
    \includegraphics[width=0.9\textwidth]{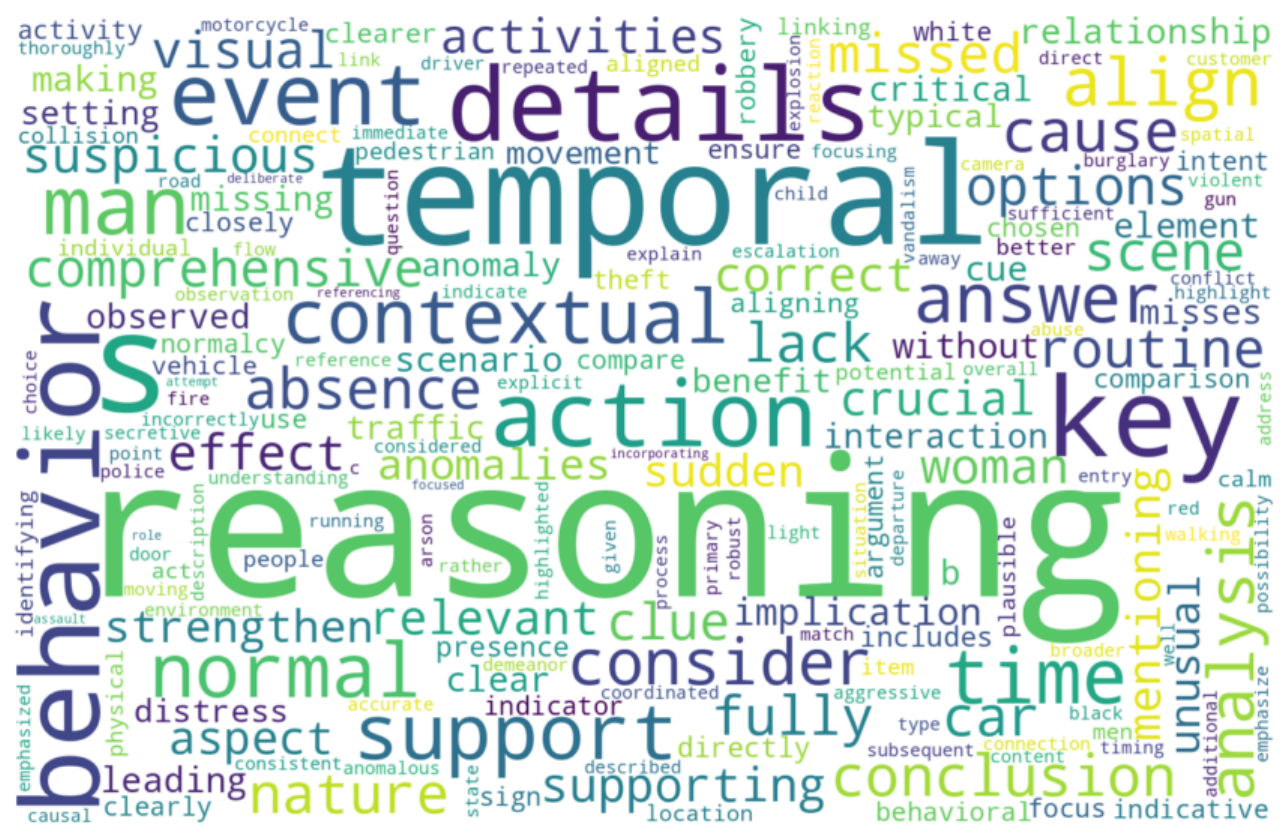}
    \caption{Word Cloud of QA Task Questions and Answer Options.}
    \label{fig:qa_wordcloud}
  \end{subfigure}
  \caption{
  \textbf{Overview of Dataset Composition and Semantic Distribution.} (a) Dataset split statistics for the QA and Temporal Grounding tasks, including train, validation, and test splits. (b) Word cloud visualization constructed from the reflection texts in the training data, highlighting dominant semantic cues learned during reflection-based anomaly reasoning.
  }
  \label{fig:dataset_and_wordcloud}
  \vskip -0.1in
\end{figure*}

Our training data are drawn from three established video anomaly datasets, UCF Crime, ECVA, and MSAD, provided by VAU-R1~\cite{zhu2025vau}. These datasets are complementary in anomaly categories, semantic complexity, and scene diversity, and together provide broad coverage for video anomaly understanding and reasoning. Detailed statistics of the datasets are reported in \cref{fig:dataset_and_wordcloud}.

\textbf{UCF-Crime} is one of the most widely used large-scale real-world datasets in video anomaly detection and understanding, mainly collected from surveillance scenarios. It covers a wide range of high-risk anomalous events, including violent behaviors, robbery, fighting, and traffic accidents. The dataset is characterized by long video durations, sparse anomaly segments, and complex backgrounds. Since anomalies typically occupy only a small temporal portion of each video, UCF-Crime poses significant challenges for temporal modeling and long-range dependency reasoning.

\textbf{ECVA} places stronger emphasis on the semantic structure and causal relations of anomalous events. It contains anomaly videos with explicit behavioral patterns and interpretable cues. Compared to detection-oriented datasets, ECVA focuses more on the process and logical structure of anomaly occurrence, providing a more suitable semantic foundation for anomaly understanding and reasoning tasks, particularly for analyzing triggering conditions and consequences of anomalous behaviors.

\textbf{MSAD} \cite{zhu2024advancing} covers anomalous events across diverse environments, including indoor, outdoor, and public spaces. This dataset exhibits high diversity in anomaly types, scene distributions, and visual appearances. Anomalies in MSAD are often more fine-grained and semantically ambiguous, posing greater challenges for model generalization and cross-scene reasoning. As such, MSAD is particularly suitable for evaluating the stability of anomaly understanding under out-of-distribution (OOD) scenarios. 

\textbf{Reflection-Enhanced Data Construction.}
To construct reflection-enhanced training data, we follow a unified ``initial reasoning → self-reflection → reasoning revision'' paradigm, explicitly modeling the complete reasoning trajectory from erroneous judgment to self-correction in video anomaly understanding. First, we employ a base multi-modal large language model, Qwen2.5-VL-3B, to generate an initial reasoning process and initial prediction for each video–question pair. At this stage, no reflection prompts are introduced intentionally, allowing the model to exhibit its natural reasoning behavior in the absence of self-reflective capability. This design provides realistic and challenging starting points for subsequent reflection. Next, based on the initial reasoning outputs and the corresponding ground-truth factual annotations, we introduce a high-capacity multi-modal teacher model, Qwen3-VL-30B, to perform systematic self-reflection over the initial reasoning. The reflection process focuses on the following aspects: (1) Whether key evidence is missing or the reasoning overly relies on local visual segments. (2) Whether temporal misalignment or errors in anomaly time localization occur. (3) Whether the reasoning chain contains causal breaks or logical jumps. (4) Whether the initial conclusion is consistent with the overall video semantics. The reflection is provided in explicit natural language, emphasizing the identification and analysis of potential issues in the initial reasoning rather than directly presenting the correct answer. Guided by the reflection feedback, the teacher model further generates a revised reasoning process and the final correct answer. The revised reasoning explicitly demonstrates how reflection influences the adjustment of the reasoning path, such as re-examining critical temporal segments, incorporating previously overlooked behavioral cues, or correcting flawed causal assumptions. This results in a logically coherent and evidence-grounded anomaly explanation. Through this procedure, reasoning failures and correction processes in video anomaly understanding are explicitly incorporated into the supervision signals. This enables the model to systematically learn how to detect errors, why errors occur, and how to correct them, providing a high-quality initialization for the subsequent reflection-aware reinforcement learning stage.

\section{Hyper-parameters}
During the reinforcement learning stage of SRVAU-R1, we adopt Group Relative Policy Optimization (GRPO) as the optimization framework. Both the rollout batch size and the training batch size are set to 128, and for each input query we sample $G=4$ candidate responses to enable group-wise relative advantage estimation, with the sampling temperature fixed to 1.0. AdamW is used as the optimizer with a learning rate of $2 \times 10^{-5}$, and all experiments are conducted using bfloat16 (bf16) mixed-precision training. During policy updates, we employ GRPO’s group-normalized advantage estimation together with a KL-divergence regularization term to constrain the distributional shift between the current policy and a frozen reference model. For computational efficiency, gradient checkpointing and FlashAttention are enabled to reduce memory consumption. All models are trained with full-parameter fine-tuning, without using LoRA or other parameter-efficient tuning modules. For video inputs, frames are uniformly sampled at 1 FPS. During reward computation, multiple reward components, including task correctness, reflection quality, and temporal IoU, are normalized within each group to reduce reward scale imbalance and stabilize multi-objective policy optimization.

\begin{table}[t]
\caption{Comparison of Performance on ECVA Datasets.}
\label{tab:ecva_rft_sft}
\centering
\setlength{\tabcolsep}{5pt}
\renewcommand{\arraystretch}{1.15}
\begin{small}
\begin{tabular}{c l cc cccccc}
\toprule
\multirow{2}{*}{Dataset} & \multirow{2}{*}{Model}
& \multicolumn{2}{c}{QA Accuracy}
& \multicolumn{6}{c}{VAU-Eval} \\
\cmidrule(lr){3-4}\cmidrule(lr){5-10}
& & Acc$_{\mathrm{w/o\ think}}$ & Acc$_{\mathrm{w/\ think}}$
& CLS$\uparrow$ & KM$\uparrow$ & FLU$\uparrow$ & INF$\uparrow$ & FAC$\uparrow$ & Total$\uparrow$ \\
\midrule
\multirow{7}{*}{ECVA}
& InternVL2.5-2B
& 78.84 & 58.84 & 2.86 & 2.78 & 7.57 & 4.62 & 3.03 & 20.86 \\
& Qwen2.5-VL-7B
& 83.02 & 86.98 & 3.70 & 3.67 & 8.64 & 6.40 & 4.04 & 26.45 \\
& InternVL2.5-8B-MPO
& 90.00 & 83.72 & 3.40 & 3.31 & 7.87 & 4.48 & 3.47 & 22.53 \\
\cmidrule(l){2-10}

& Qwen2-VL-2B
& 86.98 & 83.95 & 2.41 & 2.36 & 7.81 & 3.81 & 2.57 & 18.96 \\
& Qwen2.5-VL-3B
& 85.58 & 75.81 & 2.21 & 2.58 & 8.33 & 5.02 & 2.75 & 20.89 \\
& VAU-R1
& 89.53 & 86.51 & 1.45 & 2.24 & 8.05 & 4.32 & 2.39 & 18.45 \\
& \textcolor{red}{SRVAU-R1}
& \textbf{90.18} \textcolor{red}{(\,$\uparrow$0.65\,)}
& \textbf{92.22} \textcolor{red}{(\,$\uparrow$5.71\,)}
& 2.86 & 3.78 & 8.96 & 5.45 & 3.05 & 24.1 \\
\bottomrule
\end{tabular}
\end{small}
\end{table}
\section{Further Evaluations Results under Complex Video Scenarios (ECVA)}
\label{sec:ecva}

\cref{tab:ecva_rft_sft} reports the QA accuracy and VAU-Eval reasoning scores of different multi-modal models on the ECVA dataset. Compared with MSAD and UCF-Crime, ECVA presents greater challenges due to longer video durations, frequent viewpoint changes, and higher anomaly diversity. These factors place stronger demands on temporal understanding and causal reasoning, leading to an overall performance drop across models. In terms of QA accuracy, SRVAU-R1 achieves the best results under both w/o think and w/ think settings, indicating that reflection-aware training improves final decision-making in complex scenarios. Regarding VAU-Eval metrics, SRVAU-R1 obtains relatively higher scores on Fluency (FLU) and Informativeness (INF), suggesting better coverage of key evidence and improved logical coherence in generated reasoning. However, compared with its performance on MSAD and UCF-Crime, the overall reasoning scores of SRVAU-R1 on ECVA remain constrained. In particular, improvements on Key Matching (KM) are limited, highlighting the difficulty of long-range dependencies and scenarios with multiple co-occurring anomalies. Overall, results on ECVA indicate that reflection-enhanced training can consistently improve decision accuracy and reasoning quality in complex videos, but it is not sufficient to fully address fine-grained reasoning challenges under high temporal and semantic complexity. More targeted reflection supervision or finer-grained reward designs may be necessary for further improvements.

\section{Prompt Template}
\label{sec:prompt_template}
\begin{tcolorbox}[
  title=\textbf{Initial Reasoning Prompt},
  label={box: Initial Reasoning Prompt},
  colback=teal!6,
  colframe=teal!60,
  coltitle=black,
  fonttitle=\bfseries,
  boxrule=0.8pt,
  arc=10pt,
  left=10pt,
  right=10pt,
  top=8pt,
  bottom=8pt
]
\textbf{System Prompt:}\\
You are a precise and reliable AI assistant for video anomaly understanding.\\
You must strictly follow the formatting and reasoning rules below:
\begin{enumerate}
  \item First generate a concise, human-like reasoning process wrapped inside \texttt{<think>}...\texttt{</think>}.
  \item The reasoning must be strictly grounded in observable actions and events in the video. Do not speculate or introduce information not supported by visual evidence.
  \item Output the final answer wrapped inside \texttt{<answer>}...\texttt{</answer>}.
  \item The answer must be a single uppercase letter (A/B/C/D).
\end{enumerate}
\textbf{User:}\\
Please analyze the given video and answer the following multiple-choice question.\\
\textbf{Question:} \texttt{\{data[``Question'']\}}\\
\textbf{Options:}
\begin{itemize}
  \item \textbf{A.} \texttt{\{data[``Option 1'']\}}
  \item \textbf{B.} \texttt{\{data[``Option 2'']\}}
  \item \textbf{C.} \texttt{\{data[``Option 3'']\}}
  \item \textbf{D.} \texttt{\{data[``Option 4'']\}}
\end{itemize}
\end{tcolorbox}

\begin{tcolorbox}[
  title=\textbf{Reflection-Aware Data Construction Prompt},
  label={box: Reflection Data Construction Prompt},
  colback=teal!6,
  colframe=teal!60,
  coltitle=black,
  fonttitle=\bfseries,
  boxrule=0.8pt,
  arc=10pt,
  left=10pt,
  right=10pt,
  top=8pt,
  bottom=8pt
]
\textbf{System Prompt:}\\
You are a meticulous multi-modal reasoning editor.
You must improve a model’s reasoning by performing self-reflection and producing a revised,
grounded chain-of-thought that matches the correct option.
\vspace{0.5em}
\textbf{User:}\\
You are performing self-reflection on a previously generated reasoning process for a video-based
multiple-choice question. Your goal is to:
\begin{enumerate}
  \item Critique the initial reasoning; and
  \item Produce a cleaner, more reliable revised reasoning that supports the correct option.
\end{enumerate}
\vspace{0.5em}
\textbf{Reflection Rules:}
\begin{itemize}
  \item \textbf{If the initial answer does \emph{not} match the correct option:}
  \begin{itemize}
    \item Explain why the chosen incorrect option is not supported by the video.
    \item Explain which key evidence supports the correct option.
  \end{itemize}
  \item \textbf{If the initial answer \emph{matches} the correct option:}
  \begin{itemize}
    \item Explain how the reasoning can be clearer, better structured, or more grounded in
    chronological evidence.
  \end{itemize}
\end{itemize}
Keep the reflection short and focused on improving the reasoning structure.
\vspace{0.5em}
\textbf{Final Output Requirement:}
\begin{itemize}
  \item Generate a \textbf{FINAL THINK}:
  \item Write a new, improved reasoning wrapped inside \texttt{<think>}...\texttt{</think>}.
\end{itemize}
\end{tcolorbox}
\begin{tcolorbox}[
  title=\textbf{Self-Reflection SFT Prompt},
  label={box: Self-reflection SFT Prompt},
  colback=teal!6,
  colframe=teal!60,
  coltitle=black,
  fonttitle=\bfseries,
  boxrule=0.8pt,
  arc=10pt,
  left=10pt,
  right=10pt,
  top=8pt,
  bottom=8pt
]
\textbf{System Prompt:}\\
You are a thoughtful multi-modal reasoning model trained to analyze videos step by step
and improve your reasoning through self-reflection.
\vspace{0.5em}
\textbf{User:}\\
You are given a question about a video and four answer options (A, B, C, D).
Your goal is to perform two rounds of reasoning:
\begin{enumerate}
  \item \textbf{Initial Reasoning:} Analyze the video and generate an initial
  Chain-of-Thought (CoT) reasoning, followed by an initial answer.
  \item \textbf{Reflection and Revision:} Reflect on your initial reasoning,
  identify potential issues or areas for improvement, and then produce a revised
  reasoning and a final answer.
\end{enumerate}
\vspace{0.5em}
\textbf{Reasoning Guidelines:}
\begin{itemize}
  \item All reasoning must be strictly grounded in observable video content.
  Do not speculate or hallucinate.
  \item Follow the chronological order of events in the video.
  \item Structure reasoning as:
  \emph{what happens} $\rightarrow$ \emph{what it implies} $\rightarrow$
  \emph{comparison with options}.
  \item Keep reasoning clear, concise, factual, and well-organized.
\end{itemize}
\end{tcolorbox}

\begin{tcolorbox}[
  title=\textbf{Self-Reflection RFT Prompt},
  label={box: Self-reflection RL Prompt},
  colback=teal!6,
  colframe=teal!60,
  coltitle=black,
  fonttitle=\bfseries,
  boxrule=0.8pt,
  arc=10pt,
  left=10pt,
  right=10pt,
  top=8pt,
  bottom=8pt
]

\textbf{System Prompt:}\\
You are a multi-modal reasoning model for video understanding.
You must base your reasoning strictly on observable evidence in the video
and avoid hallucination or unsupported speculation.

\vspace{0.5em}
\textbf{User:}\\
You will be given a question about a video and four answer options (A--D).
Your task is to reason in two rounds:
\begin{enumerate}
  \item Provide an \textbf{INITIAL} reasoning and an \textbf{INITIAL} answer.
  \item Reflect on your initial reasoning, revise it if necessary,
  and provide a \textbf{FINAL} reasoning and a \textbf{FINAL} answer.
\end{enumerate}

\vspace{0.5em}
\textbf{Output Requirements (MUST follow exactly):}\\
The output must follow five parts in order:
\begin{itemize}
  \item \texttt{THINK} (\texttt{<think>}...\texttt{</think>})
  \item \texttt{ANSWER} (\texttt{<answer>}...\texttt{</answer>})
  \item \texttt{REFLECTION} (\texttt{<reflection>}...\texttt{</reflection>})
\end{itemize}

\vspace{0.5em}
\textbf{Reasoning Constraints:}
\begin{itemize}
  \item Ground all reasoning in visible events and actions in the video.
  \item Follow the chronological order of events.
  \item Keep the reasoning clear, factual, and concise.
  \item Do not introduce details not supported by the video.
\end{itemize}

\end{tcolorbox}

\section{Generated Samples Analysis}
\label{sec:generation_sample}
\begin{tcolorbox}[
  colback=cyan!6,
  colframe=cyan!35!black,
  title=\textbf{Example: Reflection-Enhanced Correction When the Initial Prediction Is Incorrect},
  fonttitle=\bfseries,
  arc=10pt,
  boxrule=0.6pt,
  left=6pt,right=6pt,top=6pt,bottom=6pt,
  enhanced,
  breakable
]

\textbf{Input Video (Sampled Frames):}

\begin{center}
\includegraphics[width=0.95\linewidth]{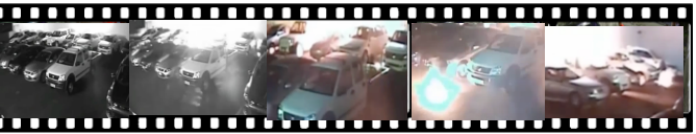}
\end{center}

\vspace{0.5em}
\textbf{Question:} What is the most critical behavioral clue that supports the conclusion of arson in the video?\\
\textbf{Options:}\\ A. The rapid spread of fire in the parking lot.\\ \textbf{\textcolor{red}{B. A man on fire is running away from the scene.}}\\ C. The presence of flammable materials in the parking lot.\\ D. The lack of firefighting equipment in the area.
\hrule
\vspace{0.4em}
\textbf{Initial Thinking (Incorrect):}

\textbf{\texttt{<think>}}
In the context of the video, the most critical behavioral clue supporting the conclusion of arson would be the rapid spread of fire in the parking lot. This suggests an intentional act rather than an accidental fire, as the fire spreads quickly without any apparent cause. \\
\textbf{\texttt{</think>}}\\
\textbf{\texttt{<answer>}}A \textbf{\texttt{<answer>}}

\textcolor{red}{\textbf{Wrong answer due to incomplete and assumption-driven reasoning.}}

\vspace{0.5em}
\hrule
\vspace{0.4em}

\textbf{Self-Reflection:}

\textbf{\texttt{<reflection>}}
The initial reasoning incorrectly focused on the rapid spread of fire but did not consider the man on fire running away, which is a more direct behavioral clue supporting arson. The expert included this detail, making the reasoning more comprehensive. Next time, ensure all key behavioral clues are considered in the analysis.
\textbf{\texttt{<reflection>}}

\vspace{0.4em}
\textbf{\texttt{</think>}}  
The video depicts two instances of intentional and deliberate acts of setting fire. First, a fire is set in a parking lot, leading to the rapid spread of flames across the area. \textcolor{red}{Oh, I see}, shortly after, a man is shown on fire and runs away from the parking lot, which suggests intentional or suspicious behavior. \textcolor{red}{Wait, let me confirm} — yes, this sequence unfolds over a very short period, with the fire spreading quickly and the man’s actions serving as strong behavioral evidence of a possible criminal act.\\
\textbf{\texttt{</think>}}\\
\textbf{\texttt{<reflection>}} B \textbf{\texttt{<reflection>}}  

\textcolor{red}{\textbf{Refinement and correction through explicit self-reflection.}}
\end{tcolorbox}
\newpage
\begin{tcolorbox}[
  colback=cyan!6,
  colframe=cyan!35!black,
  title=\textbf{Example: Reflection-Enhanced Refinement When the Initial Prediction Is Correct},
  fonttitle=\bfseries,
  arc=10pt,
  boxrule=0.6pt,
  left=6pt,right=6pt,top=6pt,bottom=6pt,
  enhanced,
  breakable
]

\textbf{Input Video (Sampled Frames):}

\begin{center}
\includegraphics[width=0.95\linewidth]{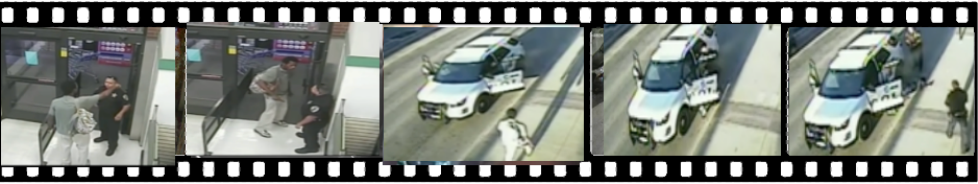}
\end{center}

\vspace{0.5em}
\textbf{Question:} What was the key action by the man in the yellow shirt that escalated the situation into an arrest anomaly?\\
\textbf{Options:}\\ A. He attempted to flee from the police officers.\\ \textbf{\textcolor{red}{B. He punched the driver of the white car, leading to a violent confrontation.}}\\ C. He verbally abused the police officers during the stop.\\ D. He ignored the police officers' commands and continued walking.

\hrule
\vspace{0.4em}
\textbf{Initial Thinking (correct):}

\textbf{\texttt{<think>}}
In the video, the man in the yellow shirt is seen approaching a white police vehicle... This action suggests that he was attempting to evade capture or escape, which is a common tactic used in situations where someone is trying to avoid arrest. The other options do not align with the described actions:- Option A: There is no....- Option C: The man...vehicle.- Option D: The man did not ... the vehicle. Therefore, the most logical conclusion is that the man's action of getting into the police vehicle was the key action that escalated the situation into an arrest anomaly\\
\textbf{\texttt{</think>}}\\
\textbf{\texttt{<answer>}} B \textbf{\texttt{<answer>}}

\textcolor{red}{\textbf{The answer is correct, but the reasoning is verbose and lacks temporal coherence.}}
\vspace{0.5em}
\hrule
\vspace{0.4em}

\textbf{Self-Reflection:}

\textbf{\texttt{<reflection>}}
The initial reasoning correctly identified the man entering the police car as a key action. However, the reasoning could be more concise and lacks temporal context and a clear representation of the actual sequence of events.
\textbf{\texttt{<reflection>}}

\vspace{0.4em}
\textbf{\texttt{</think>}}  
The video begins with a man in a yellow shirt crossing the road. A white car approaches and stops him, and two police officers exit the vehicle. \textcolor{red}{Oh, I see}, the man in yellow then walks toward the driver’s seat and punches the driver, after which he is restrained on the ground by one of the officers. \textcolor{red}{Wait, let me check} — yes, after the officer from the driver’s seat exits the car, both officers point their guns at the man lying on the ground. This sequence unfolds across several clips, clearly showing an escalation from a routine stop to a violent confrontation and arrest.\\
\textbf{\texttt{</think>}}\\
\textbf{\texttt{<reflection>}} B \textbf{\texttt{<reflection>}}  

\textcolor{red}{\textbf{After reflection, the model preserves the correct reasoning while improving linguistic coherence and temporal ordering.}}
\end{tcolorbox}

\end{document}